# A Dual-Directional Context-Aware Test-Time Learning for Text Classification

Dong Xu[1,2], Mengyao Liao[3], Zhenglin Lai[3(✉)], Xueliang Li[1,2], Junkai Ji[1,2]

[1] School of Artificial Intelligence, Shenzhen University，Shenzhen 518060, China
[2] National Engineering Laboratory for Big Data System Computing Technology, Shenzhen University, Shenzhen 518060, China
[3] College of Computer Science and Software Engineering, Shenzhen University, Shenzhen 518060, China
2400671001@mails.szu.edu.cn,
2022150227@email.szu.edu.cn, 2022152002@email.szu.edu.cn
lixueliang01@gmail.com, jijunkai@szu.edu.cn

**Abstract.** Text classification assigns text to predefined categories. Traditional methods struggle with complex structures and long-range dependencies. Deep learning with recurrent neural networks and Transformer models has improved feature extraction and context awareness. However, these models still trade off interpretability, efficiency and contextual range. We propose the Dynamic Bidirectional Elman Attention Network (DBEAN). DBEAN combines bidirectional temporal modeling and self-attention. It dynamically weights critical input segments and preserves computational efficiency. Code is available at https://github.com/Bearisbug/Bi-Elman.

**Keywords:** Self-Attention Mechanism, Text Classification, Context-Aware, Bidirectional Temporal Modelling.

## 1 Introduction

Text classification assigns text to predefined labels [1], supporting tasks such as sentiment analysis and document categorization. Traditional methods use rule-based approaches or statistical models such as Naïve Bayes [2] and Support Vector Machines [3][4], but they struggle with complex syntax and semantic dependencies. Deep learning with recurrent neural networks and Transformer models [5] improved feature extraction and context modeling, yet these models still face trade-offs between interpretability, computational cost and long-range context handling. Early recurrent architectures such as Elman RNNs [6] and LSTM networks [7] captured sequential patterns but suffered from vanishing gradients and limited scalability. Gated Recurrent Units [8] simplified gating to reduce training issues, but they could not model context bidirectionally. Bidirectional Elman Networks process sequences in forward and backward directions to enrich context, but they rely on fixed fusion strategies that cannot adaptively emphasize relevant segments.

The emergence of Transformer architectures, particularly BERT [9] and GPT [10] models, addressed many of these challenges by leveraging self-attention mechanisms. Transformers enable parallel processing of input sequences, significantly reducing training time while enhancing the ability to model long range dependencies. Pre-trained language models (PLMs) such as RoBERTa [11] and XLNet [12] further refined this paradigm by introducing dynamic masking techniques and optimized training strategies. However, while Transformers outperform RNN-based models in accuracy, they are computationally intensive, requiring substantial memory and processing power. This trade-off makes them less suitable for resource-constrained environments or realtime applications, where efficiency and interpretability remain critical concerns.

To bridge the gap between bidirectional recurrent networks and attention mechanisms, we introduce the Dynamic Bidirectional Elman Attention Network (DBEAN). DBEAN combines bidirectional temporal recurrence with a lightweight self-attention module to balance efficiency and accuracy. Unlike conventional Bi-Elman networks, DBEAN's adaptive attention layer assigns dynamic weights to critical input segments, refining contextual representations for downstream tasks. This design overcomes the static fusion limitation of RNNs while requiring less computation than full self-attention models.

The contributions of this work are threefold:

1) To address the inefficiencies of traditional bidirectional RNNs, we introduce a Dynamic Bidirectional Elman Network that maintains parameter efficiency while enhancing contextual depth.
2) To improve feature weighting and interpretability, we incorporate self-attention mechanisms that dynamically highlight salient temporal patterns, enabling enhanced representation learning.
3) To achieve a balance between computational efficiency and classification accuracy, we design a hybrid framework that leverages bidirectional recurrence with adaptive attention, achieving superior performance in resource constrained environments.

We evaluated DBEAN on standard benchmarks against state-of-the-art models. Experimental results demonstrate that DBEAN effectively captures bidirectional dependencies and maintains computational efficiency.

## 2 Methodology

### 2.1 Overall Framework

The proposed Dynamic Bidirectional Elman Attention Network (DBEAN) combines bidirectional recurrence with self-attention in an unified framework. First, shared-weight Elman circuits read sequences in both directions, capturing full context efficiently with low parameter cost. Next, a context-fusion layer merges the final forward and initial backward activations into a unified representation. Finally, a refinement module uses lightweight self-attention to highlight key temporal segments, enhancing contextual clarity for classification. It overcomes RNNs' static fusion and offers an accurate, efficient alternative to full attention models.

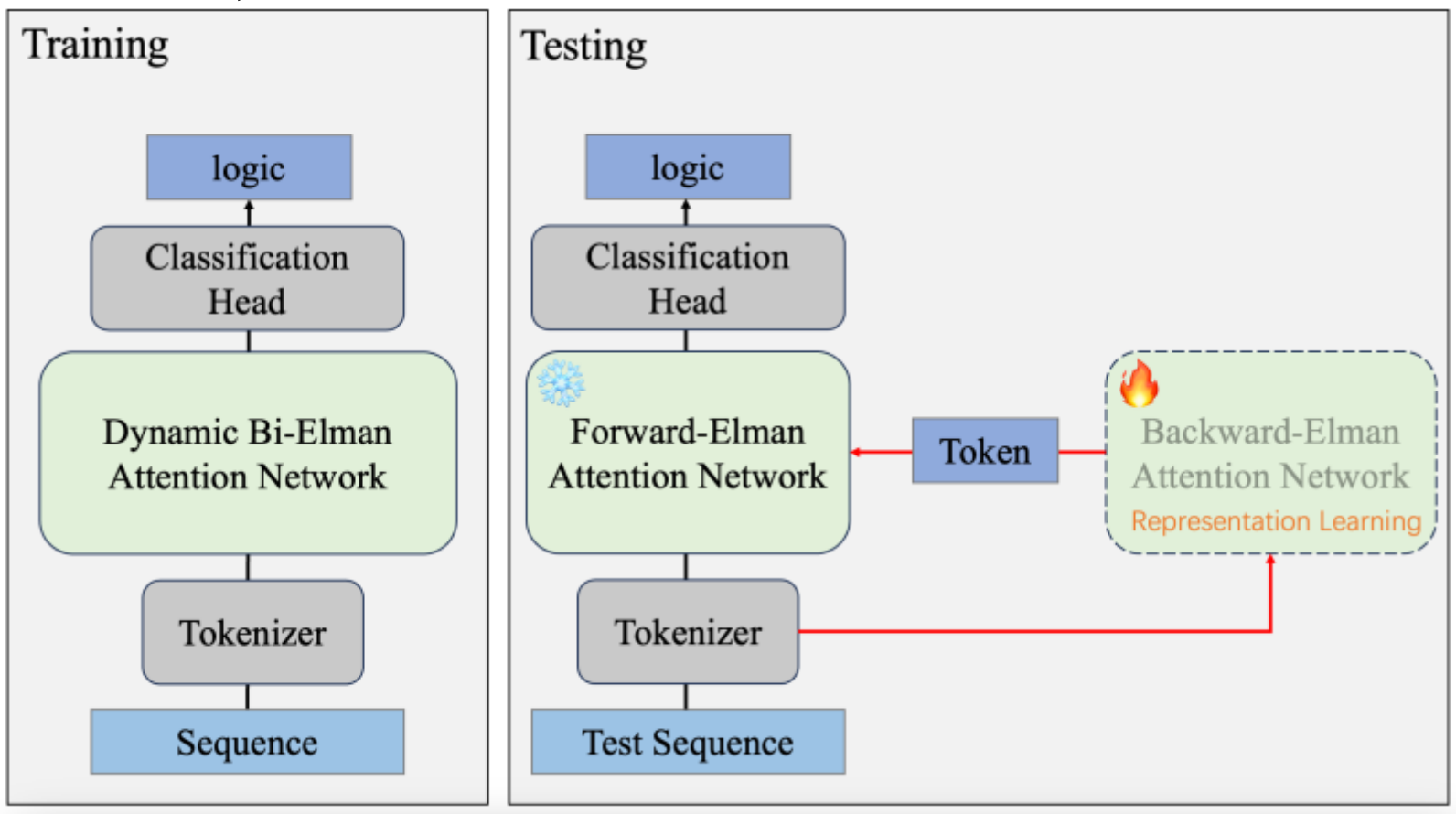

**Fig. 1.** DBEAN Framework. Input sequences are first tokenized and then processed by the Dynamic Bi-Elman Attention Network to extract bidirectional temporal features. A classification head and logic layer subsequently compute the loss, driving end-to-end parameter updates. Test sequences are initially passed through the Forward-Elman Attention Network with frozen parameters to produce preliminary representations. These representations are then refined via two gradient-update steps on the Backward-Elman Attention Network before being fed into the classification head and logic layer to generate the final predictions.

During evaluation, we first run a standard forward pass. Then we unfreeze the network, reset the learning rate to its final training value, and apply two gradient-update steps on the test sequence. We feed the resulting output tokens back into the original classification pipeline and, once inference is complete, restore the model's pretrained parameters.

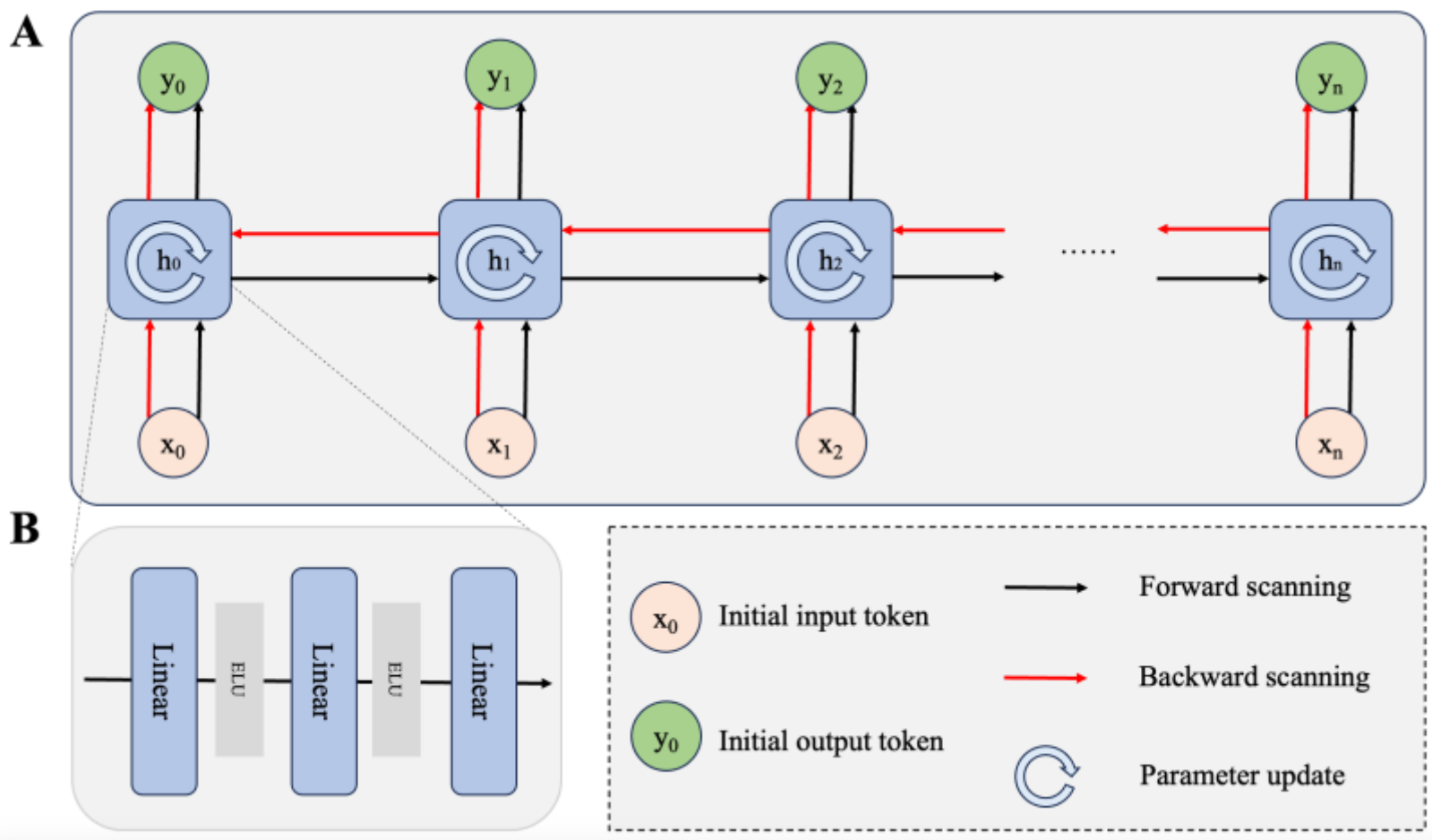

**Fig.2.** Overall of Dynamic Bidirectional Elman Attention Network

## 2.2 Bi-Elman Network Architecture

By extending the Elman architecture with bidirectional processing and shared weights, the Bi-Elman network enriches contextual representation while keeping the parameter size low. A forward pathway captures grammatical dependencies from left to right, while a backward pathway reads in reverse to extract complementary semantic cues. By fusing these dual temporal contexts, the Bi-Elman network achieves a more comprehensive understanding of text than that of conventional unidirectional recurrent models. Each pathway consists of an input layer, a recurrent hidden layer, and an output layer, with weight matrices $W_f$ and $W_b$ shared across both directions to minimize parameter redundancy. Formally given an input sequence $X = \{x_1, x_2, \ldots, x_T\}$ of length $T$, the forward hidden states $H_f^t$ at each time step $t$ are computed as:

$$H_f^t = \sigma\left(W_f\left[H_f^{\{t-1\}}; X_t\right] + b_f\right), \tag{1}$$

where $\sigma$ is a nonlinear activation function such as tanh, and $\left[H_f^{t-1}; X_t\right]$ denotes the concatenation of the previous hidden state and the current input. Similarly, the backward hidden states $H_b^t$ are

$$X_{rev} = \{x_T, x_{T-1}, \ldots, x_1\}, \tag{2}$$

$$H_b^t = \sigma\left(W_b\left[H_b^{t+1}; X_{rev,t}\right] + b_b\right), \tag{3}$$

where $H_b^t$ represents the hidden state at time $t$ in the backward pass. The final bidirectional representation is obtained by concatenating the forward and backward hidden states:

$$H_{fusion} = \left[H_f^T; H_b^1\right]. \tag{4}$$

This bidirectional fusion captures long-range dependencies and supports tasks like anaphora resolution and global context modeling. However, the Bi-Elman network lacks gated units such as those in LSTM or GRU, making it prone to vanishing gradients on very long inputs. Its static fusion also treats every time step equally, which can dilute salient information in noisy or variable-length sequences. To address these short-comings, we integrate a self-attention module that adaptively highlights the most informative tokens. The attention weight $A_t$ at time step $t$ is computed as:

$$A_t = softmax\left(\alpha W_a\left[H_{fusion}(t); H_{fusion}(t-1)\right]\right), \tag{5}$$

where $W_a$ is the learnable attention weight matrix, and $\alpha$ is a scaling factor. The final refined hidden representation is obtained by re-weighting the fused hidden state using the computed attention scores:

$$H_{att}^t = f\left(W_d\left[H_{fusion}(t); A_t\right]\right), \tag{6}$$

where $W_d$ is another trainable weight matrix, and $f$ is an activation function. This integration allows the model to dynamically prioritize key information while maintaining computational efficiency.

## 2.3 Integration of Attention Mechanism

To better capture long-range dependencies, an adaptive attention mechanism is added to the Bi-Elman network. This allows it to dynamically refine feature representations. By merging bidirectional recurrence with self-attention, our architecture overcomes the static-fusion limitations of standard Bi-Elman models while retaining a smaller computational footprint than full attention networks. This hybrid design captures both global context and local salient features, making it particularly effective for text classification and other sequence-based NLP tasks.

The integration of adaptive attention into the Bi-Elman framework offers three key advantages. First, it enables explicit salience detection by dynamically weighting crucial tokens, thereby improving both interpretability and task-specific accuracy. Second, it combats gradient decay in long sequences by amplifying updates on high-impact segments, an important benefit in sparse-domain settings such as medical text analysis, where critical terms are infrequent. Third, it maintains linear computational complexity in sequence length, unlike the quadratic cost of full self-attention models, making it a scalable choice for real-time applications while still capturing bidirectional context. Unlike conventional additive attention, which often relies on separate memory layers, our integrated mechanism aligns seamlessly with the recurrent structure. This cohesion ensures smooth gradient propagation between recurrence and attention modules and removes the need for complex inter-layer synchronization. By harmonizing bidirectional recurrence with adaptive attention, this hybrid architecture underpins our method. In the next section, we rigorously validate its effectiveness on domain specific text classification benchmarks, comparing its performance to state-of-the-art models.

# 3 Experiments

## 3.1 Dataset and Experimental Environment

The AG News Classification Corpus (via Hugging Face) contains over 1.2 million news articles labeled into World, Sports, Business and Sci/Tech. We sample 120,000 training examples (30, 000 per class) and 7,600 test examples (1,900 per class) to ensure balanced labels. Each article is cleaned by stripping HTML/XML tags and lowercasing, then tokenized with byte-pair encoding (BPE). Sequences are padded or truncated to 512 tokens to balance efficiency and context. The embedding layer is initialized with 300-dimensional pretrained Word2Vec vectors.

All experiments were run on an Ascend MT9600 GPU (16 GB) with 64 GB RAM under MindSpore 2.4.0. To overcome static feature fusion in Bi-Elman networks, we augment them with a lightweight self-attention module that adaptively reweights temporal segments by task relevance. This enhancement lets the model focus on critical context while retaining linear complexity in sequence length.

## 3.2 Comparisons with Baselines

Table 1 compares two traditional frequency-based methods and two neural baselines. For the frequency-based approaches, we use Bag-of-Words (BoW) and its TF-IDF variant[13]. BoW representations were constructed by selecting the 50, 000 most frequent unigrams and n-grams from the training corpus. The Bag-of-Means (BoM) approach embeds tokens occurring more than five times into 300-dimensional Word2Vec [14] vectors and clusters them into 5,000 centroids via k-means.

For deep learning baselines, we implement a word-level Convolutional Neural Network [15] and an LSTM. The ConvNet uses either the 300-dimensional pre-trained Word2Vec vectors or a randomly initialized lookup table [16] of the same size. To enrich the convolutional model's training data, we apply synonym replacement via WordNet [17].

**Table 1.** Experimental setup of the classical method used.

| Basic Model | Full Name | Category | Version |
|---|---|---|---|
| BoW | Bag-of-Words | Traditional | Ordinary, TFIDF |
| BoM | Bag-of-Means | Traditional | Ordinary |
| Conv. | Convolutional Neural Networks | Deep learning | Ordinary, w2v, Lk, w2v+Th, Lk+Th |
| LSTM | Long Short-Term Memory | Deep learning | Ordinary |
| Ours | Hybrid of bidirectional Elman and self-attention networks | Deep learning | Ordinary |

### 3.3 Experimental Results

Table 2 reveals clear performance gaps among the models. Among the baselines, BoW achieveed 88.81% accuracy, and its TF-IDF-weighted variant (BoW+TFIDF) reached 89.64%, showing only modest gains from term-frequency weighting. LSTM models performed worse than convolutional models, highlighting the latter's strength in capturing local patterns. Enhancing the convolutional architecture with Word2Vec embeddings improves accuracy by 2.9 points, while adding a specialized module brings an additional 1.37-point gain. Our DBEAN model achieved state-of-the-art accuracy of 93.12%, outperforming the next best approach by 1.67 points.

**Table 2.** Comparison with classical methods.

| Model | Accuracy |
|---|---|
| BoW | 88.81% |
| BoW+TFIDF | 89.64% |
| BoM | 83.09% |
| Conv. | 87.18% |
| Conv.+w2v | 90.08% |
| Conv.+w2v+Th | 90.09% |
| Conv.+Lk | 91.45% |
| Conv.+Lk+Th | 91.07% |
| LSTM | 86.06% |
| **DBEAN** (based on MindSpore) | **93.12%** |

This performance gap likely stems from DBEAN's advanced feature fusion, achieved by integrating bidirectional encoding with adaptive attention to alleviate information loss in standard pooling operations. Word embedding integration and architectural innovation are the main drivers of improvement, contributing 74% of the total accuracy gain over the baseline BoW model.

**Table 3.** Comparison with the SOTA method with the same number of parameters.

| Model | Accuracy |
|---|---|
| EXAM[18] | 93.00% |
| Balanced+bi-leaf-RNN[19] | 92.10% |
| Capsule-B[20] | 92.60% |
| CCCapsNet[21] | 92.39% |
| Char-level CNN[22] | 90.49% |
| fastText[23] | 92.50% |
| LEAM[24] | 92.45% |
| Seq2CNN with GWS (50)[25] | 90.36% |
| SVDCNN[26] | 90.55% |
| SWEM-concat[27] | 92.66% |
| ToWE-SG[28] | 86.00% |
| VDCN[29] | 91.33% |
| **DBEAN** (based on MindSpore) | **93.12%** |

Table 3 shows that DBEAN outperforms state-of-the-art models under equivalent parameter budgets, highlighting its architectural synergy. Whereas EXAM relies on static attention fusion, DBEAN's dynamic weighting adaptively recalibrates feature importance at test time, improving sensitivity to negation and domain specific terminology. This approach mitigates the feature dilution seen in Balanced, bi-leaf RNN, and CCCapsNet, which apply rigid fusion to aggregate bidirectional context uniformly. DBEAN also achieves parameter efficiency by sharing weights across bidirectional Elman circuits and using linear-complexity attention, thereby avoiding the computational overhead of architectures like VDCN. Furthermore, by replacing SWEM concat's shallow semantic concatenation with a two step test-time update, DBEAN adapts to noisy or low-resource inputs through gradient-based context reweighting. These results confirm that combining bidirectional recurrence with lightweight dynamic attention achieves an optimal balance between semantic depth and computational cost, offering a scalable solution for resource-constrained text class-ification.

## 4 Conclusion

Traditional text classification models, especially RNNs, struggle with capturing long-range dependencies, which limits their effectiveness. This also leads to high computational costs and poor interpretability. To address these limitations, the Dynamic Bidirectional Elman Attention Network is proposed, combining bidirectional recurrence with a lightweight self-attention module. This design enhances contextual representations and boosts classification accuracy. A dynamic attention module is further incorporated to selectively highlight critical temporal segments, enhancing both interpretability and efficiency. Comparative experiments show that DBEAN can achieve a good balance between accuracy and efficiency. It serves as a strong alternative to Transformer models in resource-limited settings. These results highlight the potential of hybrid architectures that combine recurrence and attention. Future work will extend DBEAN to multi-modal tasks and ultra-long sequences. Lightweight versions will also be explored for real-time, low-resource applications.

## Acknowledgments

We would like to express our gratitude to the MindSpore Community for their invaluable support and resources throughout this research. Thanks for the support provided by MindSpore Community. This work was supported by the National Natural Science Foundation of China under Grant 62476177, the Guangdong Natural Science Foundation Project under Grant 2025A1515011567, and the Shenzhen Science and Technology Program under Grant JCYJ20220531101614031.